\documentclass{article}

\usepackage{microtype}
\usepackage{graphicx}
\usepackage{subcaption}
\usepackage{booktabs} 

\usepackage{hyperref}

\usepackage[preprint]{icml2026}

\usepackage{amsmath}
\usepackage{amssymb}
\usepackage{mathtools}
\usepackage{amsthm}
\usepackage{xcolor}
\usepackage{amsfonts}
\usepackage{multirow}

\usepackage{standalone}
\usepackage{tikz}
\usetikzlibrary{positioning,decorations.pathreplacing,calc}
\usepackage{xcolor}
\usetikzlibrary{arrows.meta}

\definecolor{historygreen}{RGB}{76,175,80}
\definecolor{decodered}{RGB}{244,67,54}
\definecolor{latentorange}{RGB}{255,152,0}
\definecolor{ignoredgray}{RGB}{224,224,224}

\newif\ifshowtodos

\ifshowtodos
  \newcommand{\TODO}[1]{\textcolor{red}{\textbf{TODO:} #1}}
  \newcommand{\dfried}[1]{\textcolor{cyan}{\textbf{DF:} #1}}
  \newcommand{\ahe}[1]{\textcolor{orange}{\textbf{AH:} #1}}
\else
  \newcommand{\TODO}[1]{}   
  \newcommand{\dfried}[1]{}   
  \newcommand{\ahe}[1]{}   
\fi

\usepackage[capitalize,noabbrev]{cleveref}

\theoremstyle{plain}

\theoremstyle{definition}

\theoremstyle{remark}

\usepackage[textsize=tiny]{todonotes}

\icmltitlerunning{Reasoning with Latent Tokens in Diffusion Language Models}

\begin{document}

\twocolumn[
  \icmltitle{Reasoning with Latent Tokens in Diffusion Language Models}

\icmlsetsymbol{equaladv}{*}
\begin{icmlauthorlist}
  \icmlauthor{Andre He}{cmu}
  \icmlauthor{Sean Welleck}{cmu,equaladv}
  \icmlauthor{Daniel Fried}{cmu,equaladv}
\end{icmlauthorlist}
\icmlaffiliation{cmu}{Carnegie Mellon University, Pittsburgh, Pennsylvania, USA. \textsuperscript{*}Equal Advising}

\icmlcorrespondingauthor{Andre He}{awhe@andrew.cmu.edu}

  \icmlkeywords{Machine Learning, ICML}

  \vskip 0.3in
]



\printAffiliationsAndNotice{}  

\begin{abstract}
Discrete diffusion models have recently become competitive with autoregressive models for language modeling, even outperforming them on reasoning tasks requiring planning and global coherence, but they require more computation at inference time. We trace this trade-off to a key mechanism: diffusion models are trained to jointly predict a distribution over all unknown tokens, including those that will not actually be decoded in the current step. Ablating this joint prediction yields faster inference but degrades performance, revealing that accurate prediction at the decoded position relies on joint reasoning about the distribution of undecoded tokens.
We interpret these as \textit{latent tokens} and introduce a method for modulating their number, demonstrating empirically that this enables a smooth tradeoff between inference speed and sample quality. Furthermore, we demonstrate that latent tokens can be introduced into autoregressive models through an auxiliary multi-token prediction objective, yielding substantial improvements on the same reasoning tasks where they have traditionally struggled. Our results suggest that latent tokens, while arising naturally in diffusion, represent a general mechanism for improving performance on tasks requiring global coherence or lookahead.
\end{abstract}

\begin{figure}[!t]
    \centering
    \includegraphics[width=\linewidth]{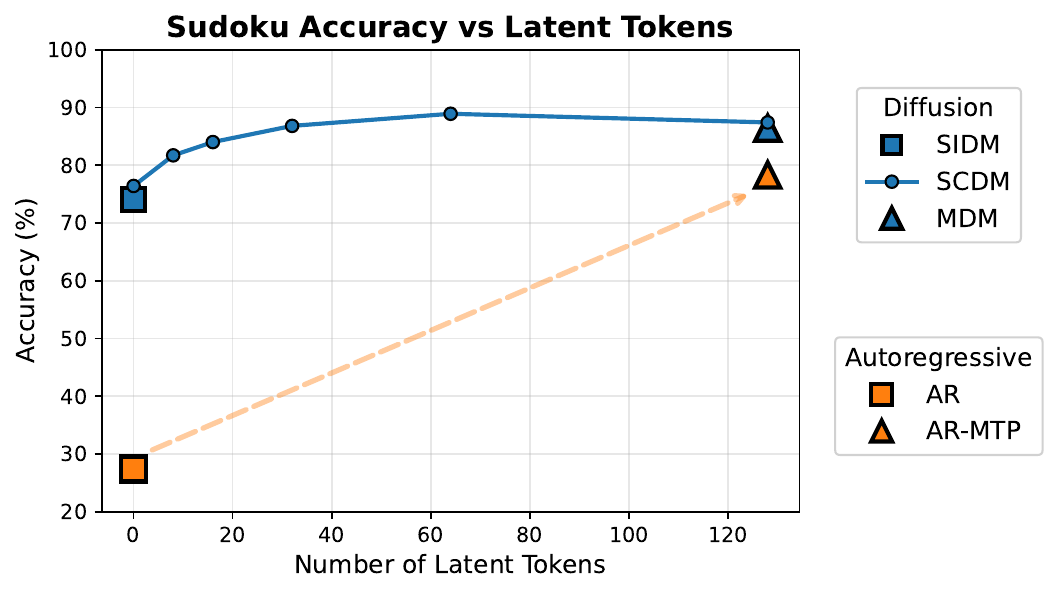}
\caption{\textbf{Latent tokens--jointly predicted at inference time but not decoded--improve reasoning performance in both diffusion and autoregressive models.} On Sudoku, accuracy increases with the number of latent tokens across both paradigms. \textit{(Blue)} A semi-causal diffusion model (SCDM) with controllable latent computation interpolates between a fast, independent-prediction model (SIDM) and an accurate, joint-prediction model (MDM). \textit{(Orange)} Standard autoregressive models (AR) struggle on constraint satisfaction tasks; equipping them with latent tokens via multi-token prediction (AR-MTP) yields dramatic improvements while remaining in the autoregressive paradigm.\vspace{-2em}}

    \label{fig:page-one}
\end{figure}

\section{Introduction}
Discrete diffusion models have recently emerged as a new paradigm for language modeling. Unlike autoregressive models, which generate tokens strictly left-to-right, diffusion models support parallel token prediction and non-monotonic generation orders. They further offer a controllable trade-off between inference speed and sample quality via the number of diffusion steps \citep{austin2023structureddenoisingdiffusionmodels, sahoo2024simpleeffectivemaskeddiffusion}. Diffusion language models have demonstrated competitive--and in some cases superior--performance relative to autoregressive models on natural language and reasoning tasks, at scales up to one hundred billion parameters \citep{ye2025dream7bdiffusionlarge,nie2025largelanguagediffusionmodels,bie2025llada20scalingdiffusionlanguage}.

Emerging evidence suggests that diffusion language models perform particularly well on tasks involving planning and constraint satisfaction \citep{ye2025dream7bdiffusionlarge, ye2025autoregressiondiscretediffusioncomplex, kim2025trainworstplanbest}. In these settings, autoregressive models can struggle to achieve comparable performance without additional scaffolding, such as training on oracle-generated solution orders.
Existing work has attributed this gap to improved subgoal decomposition \citep{ye2025autoregressiondiscretediffusioncomplex} and flexible generation order \citep{kim2025trainworstplanbest} in diffusion models; however, it remains unclear whether these factors fully account for the observed differences. 

In this work, we identify a \emph{latent token} phenomenon that helps explain the empirical advantages of diffusion language models on these tasks. In masked diffusion, generation begins from a fully masked sequence and proceeds by iteratively unmasking tokens. At each diffusion step, the model is conditioned on a partially masked sequence and \textit{jointly} predicts a distribution over tokens at all masked positions. Although only a subset of these predictions will actually be decoded (unmasked) at each step, the representations for the remaining masked-token predictions still participate since attention is bidirectional. We find that ablating this joint prediction mechanism--by training a model that predicts each masked token independently--accelerates inference but substantially degrades generation quality. 

We argue that this performance drop reveals a crucial but underappreciated role played by masked tokens in diffusion models: they act as auxiliary computational states that facilitate prediction at the target decoding position. Importantly, these auxiliary positions are themselves trained to predict their own unmasked tokens, and therefore learn to encode meaningful intermediate representations rather than merely acting as additional parameters. This is particularly useful in tasks where inferring a single unknown variable requires reasoning about many others (e.g., Sudoku). By analogy to chain of (continuous) thoughts \citep{wei2023chainofthoughtpromptingelicitsreasoning, hao2025traininglargelanguagemodels}, where latent thought tokens faciliate the final output, we refer to these auxiliary positions as \emph{latent tokens}.

This perspective motivates our central question: how does the amount of latent computation affect sequence modeling performance? We introduce \textit{latent token modulation}, a simple mechanism for controlling the number of latent tokens available during generation. 
Across both synthetic reasoning tasks and real text modeling, we evaluate models equipped with a different number of latent tokens. We find that increasing the number of latent tokens leads to consistent gains in accuracy on reasoning tasks, and consistent reduction in perplexity on language modeling. Moreover, the number of latent tokens provides a natural lever for trading off inference speed against sample quality: fewer latent tokens yield faster generation, while more latent tokens enable joint reasoning over many unknown positions.

Moving beyond masked diffusion,
we then ask whether latent tokens can be introduced into autoregressive models. We show that by augmenting autoregressive training with auxiliary multi-token prediction objectives, left-to-right models can indeed be equipped with latent tokens. On tasks where autoregressive models have previously underperformed diffusion models--often attributed to limitations of the left-to-right decomposition \citep{kim2025trainworstplanbest, ye2025autoregressiondiscretediffusioncomplex}--we find that introducing latent tokens substantially closes this gap and often surpasses diffusion models under uniform decoding. 

Our findings point to a unifying explanation for diffusion models' success on complex reasoning tasks: their ability to use latent computation from jointly predicting masked positions. By making this mechanism explicit, we identify latent tokens as a new axis, applicable in both diffusion and autoregression paradigms, to improve model performance on tasks requiring global coherence and lookahead.

\section{Masked Diffusion Models}

Discrete diffusion models \citep{austin2023structureddenoisingdiffusionmodels} have recently emerged as a competitive paradigm for language modeling, with masked diffusion \citep{sahoo2024simpleeffectivemaskeddiffusion} becoming the de facto standard \citep{nie2025largelanguagediffusionmodels,ye2025dream7bdiffusionlarge}.

We consider sequences $x = (x[1], \dots, x[L])$ of length $L$, where each token belongs to a vocabulary $\mathcal{V}$. Masked diffusion defines a forward process that progressively corrupts data over time by replacing tokens with a special \textsc{mask} token. We adopt a time indexing convention where $x_T$ denotes the clean sequence and $x_0$ denotes the fully masked sequence. At intermediate times $t \in \{1, \dots, T-1\}$, the sequence $x_t$ is partially masked: some positions contain their original clean values while others contain \textsc{mask}.

A masked diffusion model (MDM) is a bidirectional transformer trained, like a masked language model \citep{devlin-etal-2019-bert}, to predict original tokens at masked positions. Given a partially masked sequence $x_t$, the model outputs a distribution $p_\theta(x_T[\ell] \mid x_t)$ over clean tokens for each masked position $\ell$. This corresponds to learning the reverse process: progressively recovering clean tokens from masked sequences (see Appendix~\ref{app:mdm-basics} for details).

\paragraph{Inference.}
\label{sec:inference}
Samples are generated from an MDM by simulating the reverse process. Starting from the fully masked sequence $x_0$, we iteratively select positions to unmask and sample tokens from the model's predicted distributions. Once unmasked, a token remains fixed for all subsequent steps.

In the standard formulation, the positions to unmask at each step are chosen stochastically, independent of the model's predictions. This independence means we can equivalently fix the decoding schedule in advance without changing the distribution over generated sequences. For simplicity, we assume one token is generated per step so that $T = L$, deferring the general case to Appendix~\ref{app:decoding-general-case}.
Generating one token per step is also widely used in practice; generating multiple tokens per step degrades performance since the decoded tokens are conditionally independent.
This lets us represent the decoding schedule as a permutation,
\[
    \pi = (\pi_1, \dots, \pi_L),
\]
where $\pi_t \in \{1, \dots, L\}$ denotes the position to be unmasked at step $t$.\footnote{We introduce the shorthand $\pi_{<t} \coloneqq \{\pi_1, \dots, \pi_{t-1}\}$, with analogous definitions for $\pi_{\leq t}$, $\pi_{>t}$, and $\pi_{\geq t}$.}
Under this view, sampling proceeds by iterating through $t = 1, \ldots, L$: at each step, the model makes a forward pass on the current state $x_{t-1}$, computes predictions for some set of positions $P_t$, and samples a token for position $\pi_t$ from the predicted distribution. \Cref{alg:mdlm-schedule} provides pseudo-code.

\begin{algorithm}[h]
\caption{Masked diffusion sampling with schedule $\pi$}
\label{alg:mdlm-schedule}
\begin{algorithmic}[1]
\STATE \textbf{Input:} model $p_\theta$, schedule $\pi = (\pi_1, \dots, \pi_L)$
\STATE Initialize $x_0 \leftarrow [\textsc{mask}, \dots, \textsc{mask}]$
\FOR{$t = 1, 2, \dots, L$}
    \STATE Compute $p_\theta(x[\ell] \mid x_{t-1})$ jointly for $\ell \in P_t$
    \STATE Sample and set $x_{t}[\pi_t] \sim p_\theta(x[\pi_t] \mid x_{t-1})$
\ENDFOR
\STATE \textbf{Return} $x_L$
\end{algorithmic}
\end{algorithm}

\section{Prediction Sets}
\label{sec:prediction-sets}

The inference procedure specifies which token to \emph{decode} at each step ($\pi_t$), but leaves open which tokens the model \emph{predicts}. We denote this prediction set by $P_t$. In this section, we characterize this design space and describe architectures that enable different choices of $P_t$.

\subsection{Next Token Prediction vs.\ Joint Token Prediction}
\label{sec:ntp-vs-jtp}
The \emph{minimal} prediction required at step $t$ is simply
\[
    p_\theta\left( x[\pi_t] \mid x[\pi_{<t}] \right),
\]
corresponding to $P_t = \{\pi_t\}$. This predicts only the next scheduled token given all previously generated tokens, resembling next-token prediction in autoregressive models--indeed, \citet{kim2025trainworstplanbest} show that the learning problem for diffusion is equivalent to any-order autoregression. Autoregressive models can be viewed as a special case that uses a fixed schedule $\pi = (1, \dots, L)$ with this prediction set.

Masked diffusion models compute more than this. Their bidirectional architecture predicts \emph{all} masked tokens jointly in a single forward pass, with each masked position attending to both unmasked tokens and the hidden representations of other masked positions. This corresponds to computing
\[
    p_\theta\left( x[\pi_{\geq t}] \mid x[\pi_{<t}] \right),
\]
or $P_t = \pi_{\geq t}$. Only the prediction for position $\pi_t$ is decoded at each step; predictions for $\pi_{>t}$ are discarded. In \Cref{sec:latent-tokens}, we characterize these predicted-but-not-decoded tokens as \emph{latent tokens} that participate in the prediction of target tokens by providing intermediate, hidden representations.

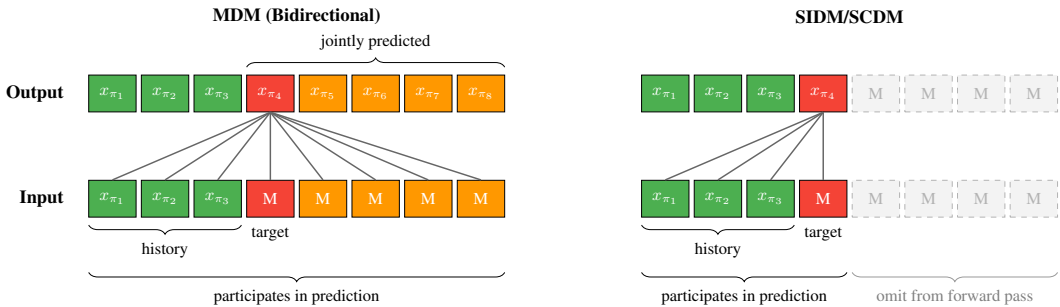
\begin{figure*}[ht]
    \centering
    \scalebox{0.7}{\begin{tikzpicture}[
    box/.style={draw, minimum width=0.9cm, minimum height=0.7cm, font=\small},
    history/.style={box, fill=historygreen, text=white},
    decode/.style={box, fill=decodered, text=white},
    latent/.style={box, fill=latentorange, text=white},
    ignored/.style={box, fill=ignoredgray, text=black!60},
    omitted/.style={box, draw=black!30, dashed, fill=ignoredgray!40, text=black!30},
    attention/.style={thick, black!60},
    rowlabel/.style={font=\bfseries, anchor=east},
    titlelabel/.style={font=\bfseries, anchor=south},
]

\def\boxsep{1.0}
\def\rowsep{2.0}
\def\figuresep{10.5}


\node[titlelabel] at ({4.5*\boxsep}, 1.2) {MDM (Bidirectional)};

\def\yout{0}

\node[history] (out1) at ({1*\boxsep}, \yout) {$x_{\pi_1}$};
\node[history] (out2) at ({2*\boxsep}, \yout) {$x_{\pi_2}$};
\node[history] (out3) at ({3*\boxsep}, \yout) {$x_{\pi_3}$};
\node[decode] (out4) at ({4*\boxsep}, \yout) {$x_{\pi_4}$};
\node[latent] (out5) at ({5*\boxsep}, \yout) {$x_{\pi_5}$};
\node[latent] (out6) at ({6*\boxsep}, \yout) {$x_{\pi_6}$};
\node[latent] (out7) at ({7*\boxsep}, \yout) {$x_{\pi_7}$};
\node[latent] (out8) at ({8*\boxsep}, \yout) {$x_{\pi_8}$};

\draw[decorate, decoration={brace, amplitude=5pt}]
    ({4*\boxsep-0.45}, \yout+0.5) -- ({8*\boxsep+0.45}, \yout+0.5)
    node[midway, above=6pt, font=\footnotesize] {jointly predicted};

\node[rowlabel] at (0.2, \yout) {Output};

\def\yin{-\rowsep}

\node[history] (in1) at ({1*\boxsep}, \yin) {$x_{\pi_1}$};
\node[history] (in2) at ({2*\boxsep}, \yin) {$x_{\pi_2}$};
\node[history] (in3) at ({3*\boxsep}, \yin) {$x_{\pi_3}$};
\node[decode] (in4) at ({4*\boxsep}, \yin) {M};
\node[latent] (in5) at ({5*\boxsep}, \yin) {M};
\node[latent] (in6) at ({6*\boxsep}, \yin) {M};
\node[latent] (in7) at ({7*\boxsep}, \yin) {M};
\node[latent] (in8) at ({8*\boxsep}, \yin) {M};

\node[rowlabel] at (0.2, \yin) {Input};

\node[font=\footnotesize, below=2pt] at (in4.south) {target};

\draw[attention] (in1.north) -- (out4.south);
\draw[attention] (in2.north) -- (out4.south);
\draw[attention] (in3.north) -- (out4.south);
\draw[attention] (in4.north) -- (out4.south);
\draw[attention] (in5.north) -- (out4.south);
\draw[attention] (in6.north) -- (out4.south);
\draw[attention] (in7.north) -- (out4.south);
\draw[attention] (in8.north) -- (out4.south);

\draw[decorate, decoration={brace, amplitude=5pt, mirror}]
    ({1*\boxsep-0.45}, \yin-0.5) -- ({3*\boxsep+0.45}, \yin-0.5)
    node[midway, below=6pt, font=\footnotesize] {history};

\draw[decorate, decoration={brace, amplitude=5pt, mirror}]
    ({1*\boxsep-0.45}, \yin-1.4) -- ({8*\boxsep+0.45}, \yin-1.4)
    node[midway, below=6pt, font=\footnotesize] {participates in prediction};


\node[titlelabel] at ({\figuresep+4.5*\boxsep}, 1.2) {SIDM/SCDM};

\node[history] (Rout1) at ({\figuresep+1*\boxsep}, \yout) {$x_{\pi_1}$};
\node[history] (Rout2) at ({\figuresep+2*\boxsep}, \yout) {$x_{\pi_2}$};
\node[history] (Rout3) at ({\figuresep+3*\boxsep}, \yout) {$x_{\pi_3}$};
\node[decode] (Rout4) at ({\figuresep+4*\boxsep}, \yout) {$x_{\pi_4}$};
\node[omitted] (Rout5) at ({\figuresep+5*\boxsep}, \yout) {M};
\node[omitted] (Rout6) at ({\figuresep+6*\boxsep}, \yout) {M};
\node[omitted] (Rout7) at ({\figuresep+7*\boxsep}, \yout) {M};
\node[omitted] (Rout8) at ({\figuresep+8*\boxsep}, \yout) {M};

\node[history] (Rin1) at ({\figuresep+1*\boxsep}, \yin) {$x_{\pi_1}$};
\node[history] (Rin2) at ({\figuresep+2*\boxsep}, \yin) {$x_{\pi_2}$};
\node[history] (Rin3) at ({\figuresep+3*\boxsep}, \yin) {$x_{\pi_3}$};
\node[decode] (Rin4) at ({\figuresep+4*\boxsep}, \yin) {M};
\node[omitted] (Rin5) at ({\figuresep+5*\boxsep}, \yin) {M};
\node[omitted] (Rin6) at ({\figuresep+6*\boxsep}, \yin) {M};
\node[omitted] (Rin7) at ({\figuresep+7*\boxsep}, \yin) {M};
\node[omitted] (Rin8) at ({\figuresep+8*\boxsep}, \yin) {M};

\node[font=\footnotesize, below=2pt] at (Rin4.south) {target};

\draw[attention] (Rin1.north) -- (Rout4.south);
\draw[attention] (Rin2.north) -- (Rout4.south);
\draw[attention] (Rin3.north) -- (Rout4.south);
\draw[attention] (Rin4.north) -- (Rout4.south);

\draw[decorate, decoration={brace, amplitude=5pt, mirror}]
    ({\figuresep+1*\boxsep-0.45}, \yin-0.5) -- ({\figuresep+3*\boxsep+0.45}, \yin-0.5)
    node[midway, below=6pt, font=\footnotesize] {history};

\draw[decorate, decoration={brace, amplitude=5pt, mirror}]
    ({\figuresep+1*\boxsep-0.45}, \yin-1.4) -- ({\figuresep+4*\boxsep+0.45}, \yin-1.4)
    node[midway, below=6pt, font=\footnotesize] {participates in prediction};

\draw[decorate, decoration={brace, amplitude=5pt, mirror}, black!40]
    ({\figuresep+5*\boxsep-0.45}, \yin-1.4) -- ({\figuresep+8*\boxsep+0.45}, \yin-1.4)
    node[midway, below=6pt, font=\footnotesize, text=black!50] {omit from forward pass};

\end{tikzpicture}}
 \caption{Comparison of attention patterns at decoding step $t=4$. In MDM (left), the target token attends to all positions, enabling joint prediction but requiring computation over the full sequence. In SIDM/SCDM (right), tokens cannot attend to later masked positions, allowing positions 5--8 to be omitted from the forward pass. SCDMs still have causal attention over positions 5--8, but these tokens are only used at training time.\vspace{-1em}} 
    \label{fig:attention-comparison}
\end{figure*}

\subsection{Attention Patterns and Prediction Sets}
\label{sec:attn-variants}

The choice of $P_t$ is determined by the model's attention pattern. In the standard bidirectional transformer, full self-attention couples the predictions of all masked positions, yielding $P_t = \pi_{\geq t}$. To enable different choices--such as $P_t = \{\pi_t\}$ for faster inference--we can replace full attention with structured patterns that control which tokens participate in predicting the target $\pi_t$.

\paragraph{Token Reordering.}
Rather than arranging tokens in their original order $1, \dots, L$, we reorder positions according to the decoding schedule $\pi$, while retaining the original positional embeddings. 
After this rearrangement, the partially decoded state at step $t$ takes the form
\[
    \tilde{x}_t = \big[\, \underbrace{x[\pi_1], \dots, x[\pi_{t-1}]}_{\text{clean tokens}},\; \underbrace{\textsc{mask}, \dots, \textsc{mask}}_{L - t + 1 \text{ masked tokens}} \,\big].
\]
This reordering is a non-functional change due to the permutation equivariance of transformers, but simplifies the description of the attention patterns below.

\paragraph{Attention Mask Notation.}
Let $A \in \{0, 1\}^{L \times L}$ denote the attention mask, where $A_{ij} = 1$ indicates that position $i$ attends to position $j$. These positions are with respect to the reordered sequence $\tilde{x}_t$, and we partition them into \textbf{clean positions} $\mathcal{C}_t = \{1, \dots, t-1\}$ (already unmasked) and \textbf{masked positions} $\mathcal{M}_t = \{t, \dots, L\}$ (still masked). 

Both attention patterns below maintain full bidirectional attention among clean tokens and allow masked tokens to attend to all clean tokens, differing only in how masked tokens interact with each other.

\textbf{Independent Attention.}
Each masked token attends only to clean tokens and itself, with no attention between distinct masked positions:
\begin{align}
\label{eqn:indep-attn}
A_{ij} = \mathbf{1}\left[ j \in \mathcal{C}_t \;\text{ or }\; i = j \right].
\end{align}
By simply making token predictions independent, this is the most straightforward way to ablate joint prediction.

\paragraph{Causal Attention.}
\label{sec:SCDM-intro}
Introduced by \citet{sahoo2025esotericlanguagemodels}, this pattern allows masked tokens to attend to preceding masked tokens in the scheduled order:
\begin{align}
\label{eqn:causal-attn}
A_{ij} = \mathbf{1}\left[ j \in \mathcal{C}_t \;\text{ or }\; (i, j \in \mathcal{M}_t \text{ and } j \leq i) \right].
\end{align}

We term a model with independent attention as a \emph{semi-independent diffusion model} (SIDM) and a model with causal attention as a \emph{semi-causal diffusion model} (SCDM). The prefix ``semi'' reflects that the attention constraints apply only among masked tokens, while clean tokens retain full bidirectional attention. Training proceeds as in standard masked diffusion; see Appendix~\ref{app:sidm-scdm-training} for details. 

We compare these models to a standard MDM in Figure~\ref{fig:attention-comparison}.
SIDM and SCDM allow independent prediction of the target token (i.e., $P_t = \pi_t$), since it does not attend to other masked positions. 
As we will explore in \Cref{sec:latent-token-modulation}, SCDMs additionally afford explicit control over the number of tokens participating in joint prediction. For now, however, we treat both SIDM and SCDM simply as approaches that remove joint prediction for faster inference -- SIDM as the straightforward solution, and SCDM as representative of existing work \citep{sahoo2025esotericlanguagemodels}.

\subsection{Do Diffusion Models Need Joint Prediction?}

Masked diffusion models have achieved strong performance on planning and constraint satisfaction tasks where autoregressive models struggle \citep{ye2025autoregressiondiscretediffusioncomplex, kim2025trainworstplanbest}. Having identified joint prediction as a key distinction from the autoregressive paradigm (\S\ref{sec:ntp-vs-jtp}), we now ask whether, and to what extent, joint prediction contributes to diffusion's success on such tasks. 

\textbf{Reasoning tasks.}
We consider three reasoning benchmarks from prior work in generative modeling: Sudoku, Zebra, and Countdown \citep{shah2024causallanguagemodelingelicit, ye2025autoregressiondiscretediffusioncomplex} (See \Cref{sec:inf-scaling-latent-tokens}).
For each task, we train and evaluate both a standard MDM and an SIDM to isolate the contribution of joint prediction. All models are evaluated under top-prob adaptive decoding \citep{kim2025trainworstplanbest}.\footnote{We explain how to make our framework compatible with adaptive decoding orders in Appendix~\ref{app:adaptive-orders}.}

Because these models are relatively small, wall-clock speedups from SIDMs over MDMs are not significant at this scale. To quantify potential efficiency gains, we instead measure the total number of tokens processed during inference, where a forward pass on a sequence of length $N$ contributes $N$ to this total. Results are presented in Table~\ref{tab:joint-token-ablation}.

MDM consistently outperforms SIDM across all three tasks, suggesting that joint prediction contributes substantially to diffusion's advantage on reasoning problems. 
\autoref{table1-with-uniform} shows results with uniform decoding, where MDM retains an edge.

\begin{table}[h]
\centering
\vspace{-0.5em}
\caption{Accuracy (\%) and relative inference cost on reasoning tasks. Cost is measured as total tokens processed, reported relative to MDM (1$\times$).}
\label{tab:joint-token-ablation}
\begin{tabular}{lccc}
\toprule
& Sudoku & Zebra & Countdown \\
\midrule
MDM  & 80.2 / 1.00$\times$ & 96.9 / 1.00$\times$ & 27.3 / 1.00$\times$ \\
SIDM & 68.1 / 0.54$\times$ & 71.5 / 0.52$\times$ & 18.4 / 0.60$\times$ \\
\bottomrule
\end{tabular}
\vspace{-0.5em}
\end{table}

\textbf{Natural language.}
We also consider diffusion models trained on OpenWebText \citep{Gokaslan2019OpenWeb}, a corpus of natural language data. Following prior work, these models are evaluated by generative perplexity.

\Cref{tab:language_results} reports results from \citet{sahoo2025esotericlanguagemodels}, who compare a standard MDM against a model highly similar to an SCDM. SCDM* corresponds to their EsoLM model, which resembles an SCDM but uses causal attention over clean tokens to enable KV caching, accounting for its substantial reduction in sampling time. This SCDM-like model achieves a generative perplexity of 48.86 compared to 36.48 for a standard MDM. Although some of this gap may be explained by the change to clean token attention, it suggests that removing joint prediction likely also degrades language modeling performance in diffusion models. 

\begin{table}[h]
\vspace{-0.5em}
    \centering
    \caption{Generative perplexity and sampling time for diffusion language models trained on OpenWebText, as reported by \citet{sahoo2025esotericlanguagemodels}. SCDM* corresponds to their EsoLM model, which resembles an SCDM but uses causal attention over clean tokens to enable KV caching.}
    \label{tab:language_results}
    \begin{tabular}{lccc}
        \toprule
        Model & Length & Gen.\ PPL ($\downarrow$) & Sampling Time (s) \\
        \midrule
        MDM & 1024 & 36.48 & 752.06 \\
        SCDM* & 1024 & 48.86 & 33.33 \\
        \bottomrule
    \end{tabular}
\vspace{-0.5em}
\end{table}

\begin{figure*}[t]
    \centering
    \scalebox{0.7}{\begin{tikzpicture}[
    box/.style={draw, minimum width=0.9cm, minimum height=0.7cm, font=\small},
    history/.style={box, fill=historygreen, text=white},
    decode/.style={box, fill=decodered, text=white},
    latent/.style={box, fill=latentorange, text=white},
    ignored/.style={box, fill=ignoredgray, text=black!60},
    poslabel/.style={font=\footnotesize},
    attention/.style={thick, black!60},
    rowlabel/.style={font=\bfseries, anchor=center},
    titlelabel/.style={font=\bfseries, anchor=south},
]

\def\boxsep{1.0}
\def\rowsep{2.0}
\def\figuresep{10.5}


\node[titlelabel] at ({4.5*\boxsep}, 1.0) {Standard Inference};

\def\yout{0}

\node[history] (out1) at ({1*\boxsep}, \yout) {$x_{\pi_1}$};
\node[history] (out2) at ({2*\boxsep}, \yout) {$x_{\pi_2}$};
\node[history] (out3) at ({3*\boxsep}, \yout) {$x_{\pi_3}$};
\node[decode] (out4) at ({4*\boxsep}, \yout) {$x_{\pi_4}$};
\node[ignored] (out5) at ({5*\boxsep}, \yout) {M};
\node[ignored] (out6) at ({6*\boxsep}, \yout) {M};
\node[ignored] (out7) at ({7*\boxsep}, \yout) {M};
\node[ignored] (out8) at ({8*\boxsep}, \yout) {M};

\node[rowlabel] at (-0.8, \yout) {Output};

\def\yin{-\rowsep}

\node[history] (in1) at ({1*\boxsep}, \yin) {$x_{\pi_1}$};
\node[history] (in2) at ({2*\boxsep}, \yin) {$x_{\pi_2}$};
\node[history] (in3) at ({3*\boxsep}, \yin) {$x_{\pi_3}$};
\node[decode] (in4) at ({4*\boxsep}, \yin) {M};
\node[ignored] (in5) at ({5*\boxsep}, \yin) {M};
\node[ignored] (in6) at ({6*\boxsep}, \yin) {M};
\node[ignored] (in7) at ({7*\boxsep}, \yin) {M};
\node[ignored] (in8) at ({8*\boxsep}, \yin) {M};

\node[rowlabel] at (-0.8, \yin) {Input};
\node[font=\footnotesize, anchor=center, align=center] at (-0.8, \yin-0.7) {positional\\embedding};

\foreach \i in {1,...,8} {
    \node[poslabel] at ({\i*\boxsep}, \yin-0.7) {$\pi_{\i}$};
}

\draw[attention] (in1.north) -- (out4.south);
\draw[attention] (in2.north) -- (out4.south);
\draw[attention] (in3.north) -- (out4.south);
\draw[attention] (in4.north) -- (out4.south);


\node[titlelabel] at ({\figuresep+4.5*\boxsep}, 1.0) {Latent Token Modulation ($n$=2)};

\node[history] (Rout1) at ({\figuresep+1*\boxsep}, \yout) {$x_{\pi_1}$};
\node[history] (Rout2) at ({\figuresep+2*\boxsep}, \yout) {$x_{\pi_2}$};
\node[history] (Rout3) at ({\figuresep+3*\boxsep}, \yout) {$x_{\pi_3}$};
\node[latent] (Rout4) at ({\figuresep+4*\boxsep}, \yout) {$x_{\pi_6}$};
\node[latent] (Rout5) at ({\figuresep+5*\boxsep}, \yout) {$x_{\pi_5}$};
\node[decode] (Rout6) at ({\figuresep+6*\boxsep}, \yout) {$x_{\pi_4}$};
\node[ignored] (Rout7) at ({\figuresep+7*\boxsep}, \yout) {M};
\node[ignored] (Rout8) at ({\figuresep+8*\boxsep}, \yout) {M};

\node[history] (Rin1) at ({\figuresep+1*\boxsep}, \yin) {$x_{\pi_1}$};
\node[history] (Rin2) at ({\figuresep+2*\boxsep}, \yin) {$x_{\pi_2}$};
\node[history] (Rin3) at ({\figuresep+3*\boxsep}, \yin) {$x_{\pi_3}$};
\node[latent] (Rin4) at ({\figuresep+4*\boxsep}, \yin) {M};
\node[latent] (Rin5) at ({\figuresep+5*\boxsep}, \yin) {M};
\node[decode] (Rin6) at ({\figuresep+6*\boxsep}, \yin) {M};
\node[ignored] (Rin7) at ({\figuresep+7*\boxsep}, \yin) {M};
\node[ignored] (Rin8) at ({\figuresep+8*\boxsep}, \yin) {M};

\node[poslabel] at ({\figuresep+1*\boxsep}, \yin-0.7) {$\pi_{1}$};
\node[poslabel] at ({\figuresep+2*\boxsep}, \yin-0.7) {$\pi_{2}$};
\node[poslabel] at ({\figuresep+3*\boxsep}, \yin-0.7) {$\pi_{3}$};
\node[poslabel, text=latentorange] at ({\figuresep+4*\boxsep}, \yin-0.7) {$\pi_{6}$};
\node[poslabel, text=latentorange] at ({\figuresep+5*\boxsep}, \yin-0.7) {$\pi_{5}$};
\node[poslabel, text=decodered] at ({\figuresep+6*\boxsep}, \yin-0.7) {$\pi_{4}$};
\node[poslabel] at ({\figuresep+7*\boxsep}, \yin-0.7) {$\pi_{7}$};
\node[poslabel] at ({\figuresep+8*\boxsep}, \yin-0.7) {$\pi_{8}$};

\draw[attention] (Rin1.north) -- (Rout6.south);
\draw[attention] (Rin2.north) -- (Rout6.south);
\draw[attention] (Rin3.north) -- (Rout6.south);
\draw[attention] (Rin4.north) -- (Rout6.south);
\draw[attention] (Rin5.north) -- (Rout6.south);
\draw[attention] (Rin6.north) -- (Rout6.south);

\end{tikzpicture}}
\caption{\textbf{Latent token modulation} with an SCDM at decoding step $t=4$. Standard inference (left) predicts $x_{\pi_4}$ using only the history tokens. With latent token modulation (right, $n$=2), we rearrange the input so that two additional masked positions ($\pi_5$, $\pi_6$) appear before the target position $\pi_4$, allowing the model to attend to more latent tokens. In both cases, we only decode the prediction $x_{\pi_4}$.\vspace{-1em}}
    \label{fig:latent-token-modulation}
\end{figure*}
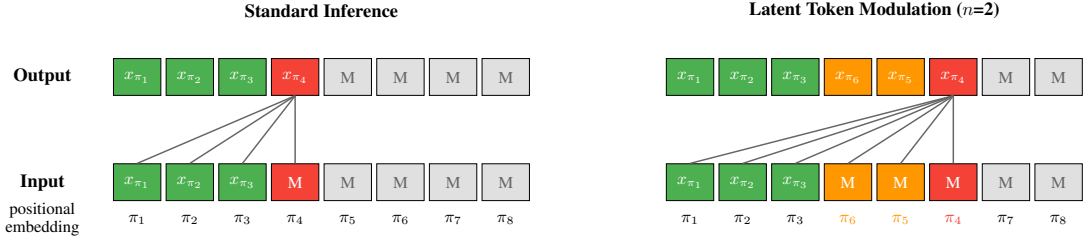

These results suggest that joint prediction plays an important role in the performance of diffusion models. In the following section, we characterize the participating masked tokens as \textit{latent tokens} and introduce a method to control their number, providing a means to trade off generation speed and quality.

\section{Latent Tokens in Diffusion Models}
\label{sec:latent-tokens}
During inference, a diffusion model may jointly predict many masked positions alongside the target position without decoding them (\S\ref{sec:ntp-vs-jtp}). These positions nonetheless contribute hidden representations that the target position can attend to, potentially serving as intermediate reasoning steps.%
\footnote{Masked positions act as both additional representational capacity and an additional source of supervision, two factors that are entangled in diffusion training. We disentangle them in \Cref{sec:ar-latent-scaling} when studying latent tokens in ARMs.}
By analogy to latent thoughts (Appendix~\ref{app:analogy-latent-reasoning}), we term these predicted-but-not-decoded tokens as \textit{latent tokens}.

\subsection{Definition}
In Sec~\ref{sec:attn-variants}, we saw that modified diffusion models can support joint prediction over different sets of positions $P_t$ during inference. Given a decoding schedule $\pi$, a time step $t$, and the prediction set $P_t$, we define the \textbf{latent tokens} as 
\(Z_t = P_t \setminus \{\pi_t\}, \)
i.e., the positions that are predicted but not decoded. We additionally require that latent tokens must participate in the computation of the target token and not be excluded by attention constraints. For example, an SIDM cannot utilize latent tokens: although it may predict multiple tokens in parallel, $\pi_t$ does not attend to any other masked positions (\autoref{eqn:indep-attn}). 

By construction, latent tokens are always masked. They share the same \textsc{mask} token embedding and are distinguished only by their positional embeddings. Thus, we characterize latent tokens solely by their indices.


\textbf{MDM.}
In a standard masked diffusion model, all masked positions participate in joint computation, so $P_t = \pi_{\geq t}$. The latent tokens at step $t$ are therefore
\(Z_t = \pi_{>t}\). This is also the \textit{maximal} set of possible latent tokens.

\textbf{SIDM.}
In a semi-independent diffusion model, masked positions cannot attend to each other, so it does not use latent tokens:
\(Z_t = \varnothing\).

\textbf{ARM.}
Left-to-right generation can be viewed as a special case with 
$\pi = (1, 2, \dots, L)$ and $P_t = \{\pi_t\}$, so autoregressive models likewise do not employ latent tokens.

We will discuss SCDMs separately below; SCDMs are particularly interesting since they will allow for a controllable number of latent tokens.

\subsection{Modulating Latent Tokens}
\label{sec:latent-token-modulation}
While our previous experiments show that joint prediction is an important mechanism in masked diffusion models (Tables~\ref{tab:joint-token-ablation} and \ref{tab:language_results}), standard masked diffusion models do not provide explicit control over the number of participating tokens: the number is determined by the context window and the number of clean tokens at the current time step. In this section, we introduce an inference algorithm for SCDMs that lets us experiment with a \textit{controllable} number of latent tokens.

Illustrated in \Cref{fig:latent-token-modulation}, the key idea is to rearrange the masked token inputs to control how many other masked positions are visible to the target under causal attention. Since masked tokens later in the input attend to more preceding masks, and diffusion models support any token order $\pi$, we can deliberately reposition other masked tokens \textit{before} the target position to introduce more latent computation. We call this \textbf{latent token modulation}. Importantly, latent token modulation \textit{does not require retraining} the model. 

We illustrate an inference step using latent token modulation in \Cref{fig:latent-token-modulation}. For a desired number of latent tokens $n$, we sample $n$ future masked positions $Z_t$ and place them before the target position $\pi_t$ in the input. While $n$ can technically vary per time step, we keep it as a single hyperparameter for simplicity. Intuitively, we are prompting the model to reason about certain other unknown positions $Z_t$ before predicting the target unknown position $\pi_t$. In the Appendix, we provide pseudocode (App.~\ref{app:latent-token-modulation}) and explain how we make this compatible with adaptive decoding orders (App.~\ref{app:adaptive-with-ltm}).

\subsection{Experiment: Latent Token Scaling}
\label{sec:inf-scaling-latent-tokens}

To quantify the contribution of latent tokens to diffusion model performance, we train SCDMs on reasoning tasks and evaluate their performance when generating samples with varying numbers of latent tokens. Additionally, we apply latent token modulation to the SCDM-like model from \citet{sahoo2025esotericlanguagemodels} to investigate the relationship between sample quality and latent token usage in language modeling.

\subsubsection{Datasets\footnote{More dataset details can be found in Appendix~\ref{app:datasets}.}}
\label{sec:reasoning-datasets}

\textbf{Sudoku-Generative}: Dataset of valid $9 \times 9$ Sudoku boards used for unconditional generation, where we evaluate the percentage of generated samples that are valid.

\textbf{Sudoku-Puzzle} \citet{shah2024causallanguagemodelingelicit}: Dataset of partially filled Sudoku puzzles used for conditional generation. The model must complete the board given revealed cells.

\textbf{Zebra} \citep{shah2024causallanguagemodelingelicit}: Also known as Einstein puzzles, these logical puzzles require assigning attributes to entities while satisfying a list of constraints.

\textbf{Countdown} \citet{ye2025dream7bdiffusionlarge}: Arithmetic game to compute a target number using 5 given numbers.

\begin{figure}[t]
    \centering
    \includegraphics[width=0.96\linewidth]{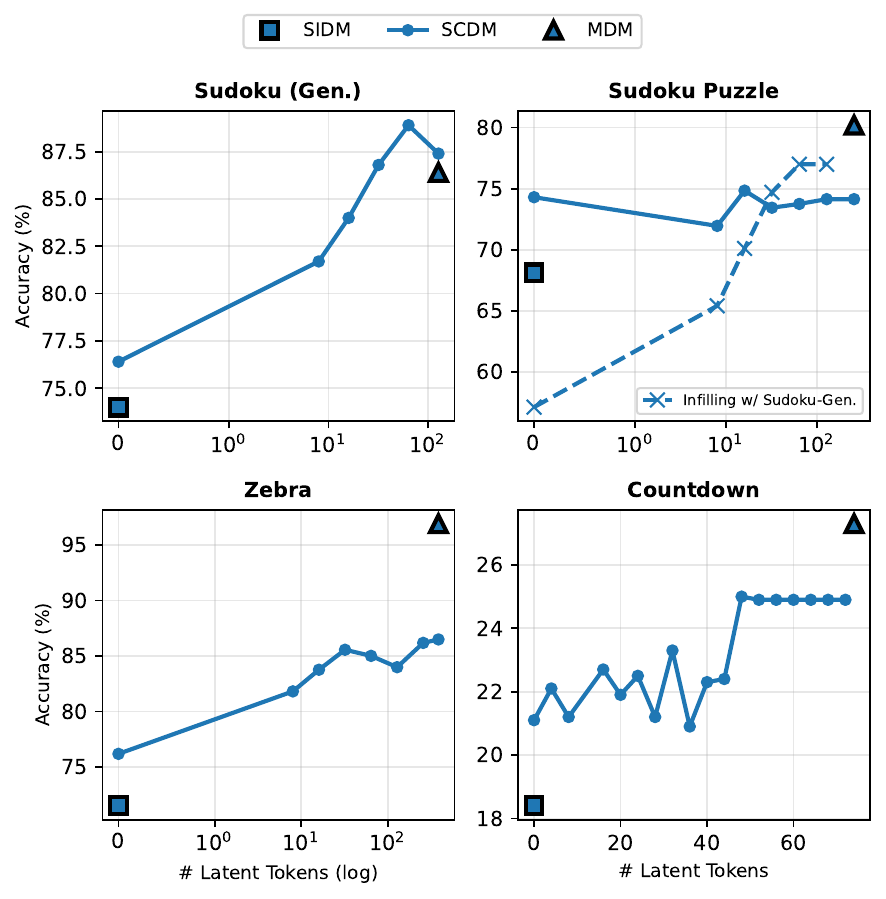}
    \vspace{-0.5em}
    \caption{Performance of a SCDM on four reasoning tasks as a function of latent token count. Under most settings, performance improves as more latent tokens are used for prediction. Varying the number of latent tokens has the effect of interpolating between SIDM ($n=0$), a fast model using no latent tokens, and MDM ($n=L$), a stronger model using all available latent tokens.}
    \label{fig:latent-tokens-four-datasets}
    \vspace{-2em}
\end{figure}

\subsubsection{Results}
\textbf{Reasoning tasks.}
For each task, we train a SCDM and evaluate it across a range of latent token counts $n$, presenting results in Figure~\ref{fig:latent-tokens-four-datasets}. Because SIDM and MDM correspond to $n=0$ and $n=L$ latent tokens respectively, we include them as reference points. Across most settings, increasing the number of latent tokens yields consistent improvements in task accuracy.

For the Sudoku Puzzle task, we additionally evaluate the model trained on \textbf{Sudoku (Generative)} by using it to infill the missing cells. Interestingly, while the model trained directly on Sudoku Puzzle data does not exhibit a strong scaling trend with latent tokens, the model trained on unconditional generation does when evaluated on the conditional task. We hypothesize that this discrepancy arises from differences in the training data distribution, and leave as future work to investigate when latent token scaling emerges. 

\begin{figure}[t]
    \centering
    \includegraphics[width=0.88\linewidth]{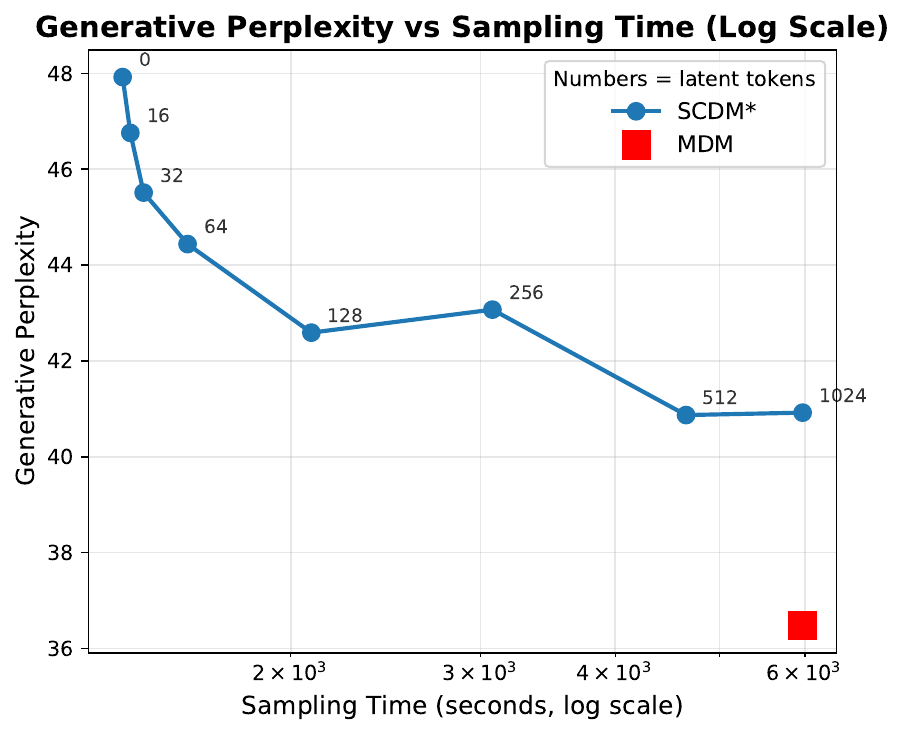}
    \caption{Generative perplexity of an SCDM-like model from \citet{sahoo2025esotericlanguagemodels} as a function of latent token count. Increasing the number of latent tokens provides a tunable trade-off between sampling speed and sample quality.}
    \label{fig:latent-tokens-owt}
    \vspace{-2em}
\end{figure}

\textbf{Natural language.}
We evaluate the SCDM-like model trained by \citet{sahoo2025esotericlanguagemodels} on OpenWebText \citep{Gokaslan2019OpenWeb}, using latent token modulation across a range of token counts. As shown in Figure~\ref{fig:latent-tokens-owt}, generative perplexity decreases nearly monotonically with additional latent tokens. By simply adding more latent tokens at inference time, we reduce generative perplexity from approximately 48 to 40 for the same trained model, substantially closing the gap to a standard MDM at 36.48.

Across both reasoning and language modeling, controlling the number of latent tokens provides a mechanism to interpolate between a fast model that forgoes joint prediction (SIDM) and a high-quality model that leverages all latent tokens (MDM). This suggests that latent token usage likely explains the performance gap between models that predict only the next token (SIDM) and those that jointly predict many unknown tokens with attention between them (MDM). On most tasks, a gap persists between SCDM with $n=0$ and SIDM, as well as between SCDM with $n=L$ and MDM--we attribute this to differences in expressivity and training dynamics arising from the modified attention structure on masked tokens. 

\section{Latent Tokens in Autoregressive Models}
\label{sec:latent-ar}
Latent tokens arise naturally in diffusion models, since they must simultaneously capture the distribution of several possible next tokens. Our results in the last section suggest that latent tokens are central to the reasoning performance of diffusion models. We now ask whether we can equip autoregressive models (ARMs) with latent tokens -- can we close the gap between ARMs and MDMs on reasoning tasks~\citep{ye2025autoregressiondiscretediffusioncomplex}
by introducing latent tokens?\footnote{Existing work introduces additional latent computation in ARMs \citet{goyal2024thinkspeaktraininglanguage}. Our framework differs in that latent tokens are specifically trained to carry predictive information about non-target tokens, rather than just being additional compute.}

\begin{figure*}[t]
    \centering
    \scalebox{0.7}{\begin{tikzpicture}[
    box/.style={draw, minimum width=0.9cm, minimum height=0.7cm, font=\small},
    history/.style={box, fill=historygreen, text=white},
    decode/.style={box, fill=decodered, text=white},
    latent/.style={box, fill=latentorange, text=white},
    ignored/.style={box, fill=ignoredgray, text=black!60},
    omitted/.style={box, draw=black!30, dashed, fill=ignoredgray!40, text=black!30},
    poslabel/.style={font=\footnotesize},
    attention/.style={thick, black!60},
    rowlabel/.style={font=\bfseries, anchor=east},
    titlelabel/.style={font=\bfseries, anchor=south},
]

\def\boxsep{1.0}
\def\rowsep{2.0}
\def\figuresep{10.5}


\node[titlelabel] at ({4.5*\boxsep}, 1.0) {Standard ARM};

\def\yout{0}

\node[history] (out1) at ({1*\boxsep}, \yout) {$x_2$};
\node[history] (out2) at ({2*\boxsep}, \yout) {$x_3$};
\node[decode] (out3) at ({3*\boxsep}, \yout) {$x_4$};
\node[omitted] (out4) at ({4*\boxsep}, \yout) {M};
\node[omitted] (out5) at ({5*\boxsep}, \yout) {M};
\node[omitted] (out6) at ({6*\boxsep}, \yout) {M};
\node[omitted] (out7) at ({7*\boxsep}, \yout) {M};
\node[omitted] (out8) at ({8*\boxsep}, \yout) {M};

\node[rowlabel] at (0.2, \yout) {Output};

\def\yin{-\rowsep}

\node[history] (in1) at ({1*\boxsep}, \yin) {$x_1$};
\node[history] (in2) at ({2*\boxsep}, \yin) {$x_2$};
\node[history] (in3) at ({3*\boxsep}, \yin) {$x_3$};
\node[omitted] (in4) at ({4*\boxsep}, \yin) {M};
\node[omitted] (in5) at ({5*\boxsep}, \yin) {M};
\node[omitted] (in6) at ({6*\boxsep}, \yin) {M};
\node[omitted] (in7) at ({7*\boxsep}, \yin) {M};
\node[omitted] (in8) at ({8*\boxsep}, \yin) {M};

\node[rowlabel] at (0.2, \yin) {Input};

\foreach \i in {1,...,8} {
    \node[poslabel] at ({\i*\boxsep}, \yin-0.7) {\i};
}

\draw[-{Stealth[length=2.5mm]}, thick] ({3*\boxsep+0.35}, \yout+0.65) -- ({1*\boxsep-0.35}, \yout+0.65);
\node[font=\footnotesize, above] at ({2*\boxsep}, \yout+0.65) {shifted};

\draw[attention] (in1.north) -- (out3.south);
\draw[attention] (in2.north) -- (out3.south);
\draw[attention] (in3.north) -- (out3.south);


\node[titlelabel] at ({\figuresep+4.5*\boxsep}, 1.0) {AR-MTP};

\node[history] (Rout1) at ({\figuresep+1*\boxsep}, \yout) {$x_1$};
\node[history] (Rout2) at ({\figuresep+2*\boxsep}, \yout) {$x_2$};
\node[history] (Rout3) at ({\figuresep+3*\boxsep}, \yout) {$x_3$};
\node[decode] (Rout4) at ({\figuresep+4*\boxsep}, \yout) {$x_4$};
\node[latent] (Rout5) at ({\figuresep+5*\boxsep}, \yout) {$x_n$};
\node[latent] (Rout6) at ({\figuresep+6*\boxsep}, \yout) {$x_n$};
\node[latent] (Rout7) at ({\figuresep+7*\boxsep}, \yout) {$x_n$};
\node[latent] (Rout8) at ({\figuresep+8*\boxsep}, \yout) {$x_n$};

\node[history] (Rin1) at ({\figuresep+1*\boxsep}, \yin) {$x_1$};
\node[history] (Rin2) at ({\figuresep+2*\boxsep}, \yin) {$x_2$};
\node[history] (Rin3) at ({\figuresep+3*\boxsep}, \yin) {$x_3$};
\node[decode] (Rin4) at ({\figuresep+4*\boxsep}, \yin) {M};
\node[latent] (Rin5) at ({\figuresep+5*\boxsep}, \yin) {M};
\node[latent] (Rin6) at ({\figuresep+6*\boxsep}, \yin) {M};
\node[latent] (Rin7) at ({\figuresep+7*\boxsep}, \yin) {M};
\node[latent] (Rin8) at ({\figuresep+8*\boxsep}, \yin) {M};

\foreach \i in {1,...,8} {
    \node[poslabel] at ({\figuresep+\i*\boxsep}, \yin-0.7) {\i};
}

\draw[attention] (Rin1.north) -- (Rout4.south);
\draw[attention] (Rin2.north) -- (Rout4.south);
\draw[attention] (Rin3.north) -- (Rout4.south);
\draw[attention] (Rin4.north) -- (Rout4.south);
\draw[attention] (Rin5.north) -- (Rout4.south);
\draw[attention] (Rin6.north) -- (Rout4.south);
\draw[attention] (Rin7.north) -- (Rout4.south);
\draw[attention] (Rin8.north) -- (Rout4.south);

\end{tikzpicture}}
    \caption{\textbf{Comparison of Standard ARM and AR-MTP architectures.} \emph{Left:} In a standard autoregressive model, outputs are shifted left relative to inputs, and causal attention restricts each position to attend only to previous positions. \emph{Right:} AR-MTP removes both the shift and the causal mask, using bidirectional attention so that each position attends to all inputs, including masked future tokens. This is architecturally equivalent to an MDM (\Cref{fig:attention-comparison}) with a fixed left-to-right generation order.}
    \label{fig:ar-mtp}
    \vspace{-1.5em}
\end{figure*}
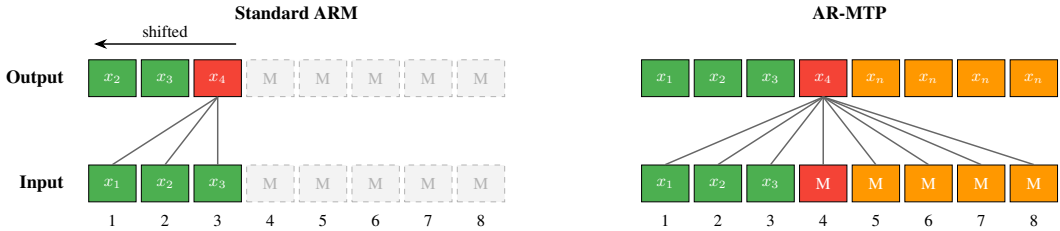

\subsection{Latent Tokens via Multi-Token Prediction}

We saw that autoregressive generation is a special case of \Cref{alg:mdlm-schedule} with a fixed left-to-right order (\Cref{sec:ntp-vs-jtp}). This suggests a natural way to introduce latent tokens into autoregressive models: train them to jointly predict multiple unknown (future) tokens, not just the immediate next one. 

To enable such multi-token prediction, we modify the standard autoregressive architecture in two ways: we remove the causal attention mask, allowing each position to attend to all inputs, and we remove the output shift, so that position $t$ predicts token $t$ rather than token $t+1$. The result, illustrated in \Cref{fig:ar-mtp}, is a bidirectional transformer that takes partially masked sequences as input and jointly predicts all masked positions. We call this model \textbf{AR-MTP}.

For training, we sample a time step $t$ for each example and construct a sequence by masking out tokens from the right:
\begin{align*}
    \tilde{x}_t = \big[\, x[1], \dots, x[t-1],\; \textsc{mask}_t, \dots, \textsc{mask}_L \,\big].
\end{align*}
The model is trained to predict all masked tokens, i.e., to model $p_\theta(x[\geq t] \mid \tilde{x}_t)$. Unlike standard ARMs, which train on all positions in parallel via causal masking, AR-MTP samples a single time step per example, similar to MDM training. See \Cref{app:ar-mtp-training} for the full objective.

At inference time, AR-MTP generates tokens autoregressively in left-to-right order, decoding only from the next token position $t$ and discarding other predictions. This means that AR-MTP still falls under the autoregressive paradigm, using multi-token prediction only as an auxiliary prediction task. AR-MTP is closely related to \citet{gloeckle2024betterfasterlarge}; we use it here not as a novel architecture, but as a tool to study how ARMs may benefit from latent tokens.

\subsection{Experiment: Reasoning Tasks}

We evaluate AR-MTP against standard ARMs on four reasoning tasks: Sudoku-generative, Sudoku-puzzle, Zebra, and Countdown (\Cref{sec:reasoning-datasets}). Results are shown in \Cref{tab:ar-mtp-reasoning-tasks}, with MDM results under uniform and top-prob decoding included for reference. \textbf{Adding latent tokens substantially improves performance across all four tasks}, even when remaining within the autoregressive framework. AR-MTP outperforms standard AR by a wide margin on every benchmark. 

\begin{table}[h]
\centering
\begin{tabular}{@{}lcccc@{}}
\toprule
Model & Sdk-G & Sdk-P & Zebra & CD \\
\midrule
ARM & 27.4 & 5.5 & 54.7 & 2.6 \\
AR-MTP & 78.3 & 53.6 & 68.8 & 8.6 \\
MDM-uniform & 40.5 & 12.1 & 26.6 & 17.9 \\
MDM-top-prob & 86.4 & 80.2 & 96.9 & 27.3 \\
\bottomrule
\end{tabular}
\caption{Accuracy (\%) on reasoning tasks. AR-MTP substantially outperforms standard AR across all tasks, indicating that latent tokens benefits autoregressive models on such tasks.}
\label{tab:ar-mtp-reasoning-tasks}
\vspace{-1em}
\end{table}

While MDM with top-prob decoding remains the strongest method overall, AR-MTP actually surpasses MDM with uniform decoding on three of four tasks. This suggests that \textit{when controlling for latent computation}, diffusion is not inherently superior to autoregression for learning these distributions. Rather, the advantage of MDMs on these tasks appears to stem primarily from their ability to use adaptive decoding orders \citep{kim2025trainworstplanbest}.

\subsection{Experiment: Latent Token Scaling}
\label{sec:ar-latent-scaling}

We next investigate whether AR-MTP performance improves with additional latent tokens. To do so, we consider a more challenging modeling task involving longer sequences and complex structural constraints.

\textbf{Sudoku-Large.} We construct a synthetic dataset of $36 \times 36$ sudoku boards, where each cell contains a digit from 1 to 36, yielding sequences of approximately 1.3K tokens. Valid boards must satisfy constraints analogous to standard sudoku: each row, column, and $6 \times 6$ block must contain exactly the numbers 1 through 36. Generating fully valid boards proves difficult for all models, so we design a task-specific metric to measure the degree of constraint satisfaction in generated samples. Details are provided in Appendix~\ref{app:sdk-large-task-loss}.

\textbf{AR-MTP-Windowed.} To vary the number of latent tokens available to AR-MTP, we train windowed variants with fixed latent token counts. As illustrated in Figure~\ref{fig:ar-mtp-win} (in Appendix), AR-MTP-Window-$n$ jointly predicts up to $n$ masked tokens while ignoring all subsequent tokens. This architecture can be viewed as using $n-1$ latent tokens.

\textbf{Ablation: AR-NTP.} To isolate the effect of the multi-token prediction objective, we consider AR-NTP: an ablation that shares the same architecture as AR-MTP-Windowed but \textit{applies the training loss only to the next token}. The remaining $n-1$ positions still participate in the forward pass, providing additional parameters and computation, but receive no gradient signal. This allows us to disentangle the benefits of multi-token training from those of increased model capacity.

\begin{figure}[t]
    \centering
    \includegraphics[width=0.9\linewidth]{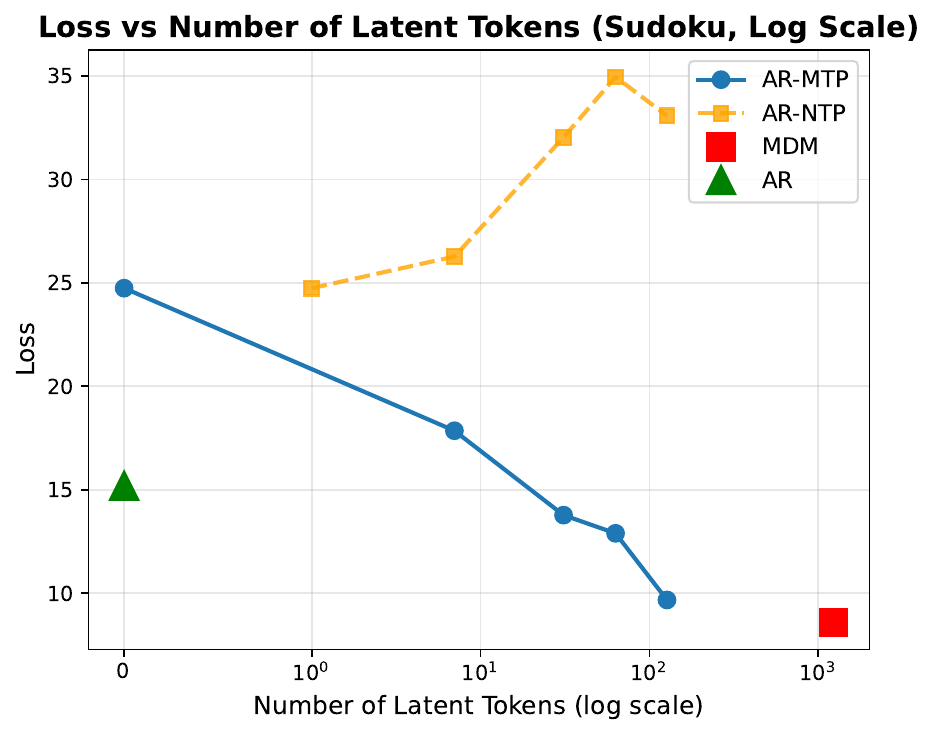}
    \caption{Sample quality on Sudoku-Large as a function of latent token count. AR-MTP (blue) improves steadily with more latent tokens, approaching MDM performance. AR-NTP (orange), which uses the same architecture but trains only the next-token prediction, degrades with additional capacity. Standard AR (green) and MDM (red) are shown for reference.\vspace{-2em}}
    \label{fig:sudoku-large-plot}
\end{figure}

\textbf{Results.} We train AR-MTP and AR-NTP models across a range of window sizes and evaluate sample quality; results are shown in Figure~\ref{fig:sudoku-large-plot}. 
AR-MTP exhibits consistent improvement as the number of latent tokens increases, with performance approaching that of MDM at $n=128$. This confirms that the latent token scaling behavior observed in diffusion models transfers to the autoregressive setting. An exception is standard ARM, which outperforms AR-MTP-Window-8 despite using no latent tokens. We attribute this gap to the benefits of timestep parallel training, a unique advantage of standard ARMs.

In contrast, AR-NTP performance \emph{degrades} as the window size increases. This suggests that the benefits of latent tokens arise specifically from training them to carry predictive information about other tokens--merely providing additional scratch space or parameters is insufficient.

\section{Related Works}

\textbf{Diffusion Language Models.} 
Diffusion models were originally developed for continuous data \citep{sohldickstein2015deepunsupervisedlearningusing}. \citet{austin2023structureddenoisingdiffusionmodels} introduced a framework for discrete diffusion, and masked diffusion has proven particularly successful for language modeling \citep{sahoo2024simpleeffectivemaskeddiffusion}. Recent work has scaled this approach to billions of parameters, achieving performance competitive with autoregressive models \citep{nie2025largelanguagediffusionmodels,ye2025dream7bdiffusionlarge,bie2025llada20scalingdiffusionlanguage}. Our semi-causal diffusion model builds on \citet{sahoo2025esotericlanguagemodels}, who introduced this attention pattern for efficient inference.

\textbf{Diffusion vs.\ Autoregressive Models.}
Several works have sought to understand when and why diffusion models outperform autoregressive models (ARMs). \citet{ye2025autoregressiondiscretediffusioncomplex} find that diffusion models excel on complex reasoning and planning tasks, attributing this to improved subgoal learning. \citet{kim2025trainworstplanbest} demonstrate that adaptive generation orders account for substantial gains over left-to-right decoding. \citet{prabhudesai2025diffusionbeatsautoregressivedataconstrained} show that diffusion models outperform ARMs in data-constrained settings. A related line of work establishes theoretical connections between MDMs and any-order autoregressive models \citep{shih2022traininginferenceanyorderautoregressive,kim2025trainworstplanbest}.

\textbf{Latent Computation.}
Several works have explored adding latent computation to language models. \citet{goyal2024thinkspeaktraininglanguage} introduce learnable pause tokens that provide additional computation before producing each output token. \citet{zelikman2024quietstarlanguagemodelsteach} train models to generate internal rationales prior to prediction. \citet{hao2025traininglargelanguagemodels} represent chain-of-thought reasoning entirely in latent space using continuous tokens. Our notion of latent tokens is related but distinct: we do not introduce a new mechanism but study latent computation that arises naturally from joint prediction over multiple unknown positions.

\section{Conclusion}
We have identified latent tokens as a key mechanism underlying the strong reasoning performance of diffusion language models. By introducing latent token modulation, we demonstrated that the number of latent tokens provides an axis for trading off inference speed against sample quality, with consistent improvements observed as more latent tokens participate in prediction. 

We also showed that latent tokens are not exclusive to diffusion: augmenting autoregressive models with multi-token prediction introduces the same mechanism and yields substantial gains on constraint satisfaction tasks where standard ARMs have struggled. 

Our findings point toward a unified view of latent computation in sequence models. In both diffusion and autoregressive paradigms, training models to jointly predict unknown tokens improves their ability to solve reasoning tasks that require planning and constraint satisfaction. 

\section{Acknowledgments}
We thank Saujas Vaduguru for helpful comments on the draft. This work was supported in part by the National Science Foundation under Grant Nos. DMS-2502281 and DMS-2434614. 

\section{Impact Statement}
This paper presents work whose goal is to advance the field of machine learning. There are many potential societal consequences of our work, none of which we feel must be specifically highlighted here.

\bibliography{example_paper}
\bibliographystyle{icml2026}

\newpage
\twocolumn
\appendix

\section{Masked Diffusion Models}
\label{app:mdm-basics}

Diffusion models \citep{sohldickstein2015deepunsupervisedlearningusing} are characterized by a forward noising process that progressively corrupts data into noise, together with a learned reverse process that transforms noise back into data. While originally developed for continuous data, \citet{austin2023structureddenoisingdiffusionmodels} introduced a framework of diffusion models for discrete data, where masked diffusion is a widely adopted variant~\cite{sahoo2024simpleeffectivemaskeddiffusion}. 

We consider sequences $x = (x^{(1)}, \dots, x^{(L)})$ of length $L$, where each token $x^{(\ell)}$ belongs to a vocabulary $\mathcal{V}$. In masked diffusion, the forward noising process progressively replaces tokens with a special $\textsc{mask}$ token. We adopt a time indexing convention where $x_0$ denotes the fully masked sequence and $x_T$ denotes the unmasked, or \emph{clean}, sequence, so that the forward process generates a trajectory of increasingly masked states $x_T, x_{T-1}, \dots, x_0$ over $T$ steps.

Diffusion models are trained to approximate the reverse transitions $p_\theta(x_{t+1} \mid x_t)$, which amounts to predicting the original clean values of masked tokens given a partially masked sequence $x_t$. To generate samples from a model, we start from a fully masked sequence and iteratively sample from the predicted token distributions to transform $x_0$ back into $x_T$.
A defining property of masked diffusion is that in the forward process, tokens no longer change after becoming \textsc{mask}; correspondingly, in the reverse process, tokens that are unmasked remain fixed for all subsequent time steps.  For this reason, masked diffusion is also known as absorbing state diffusion. This approach has proven particularly successful and become the de-facto standard for diffusion language modeling \citep{sahoo2024simpleeffectivemaskeddiffusion}.

\textbf{Model.}
Masked diffusion language models parameterize the reverse process using a masked language model (MLM) architecture. Given a partially masked sequence $x_t$, a bidirectional transformer is trained to predict the original token values at each masked position. The model takes $x_t$ as input and outputs a distribution $p_\theta(x_T[\ell] \mid x_t)$ over clean tokens for each masked position $\ell$. During inference, these predicted distributions are used to sample tokens that progressively unmask the sequence.

\textbf{Training}.
Masked diffusion models are typically trained with a variational objective that lower bounds data likelihood. \citet{sahoo2024simpleeffectivemaskeddiffusion} show that for the MLM parameterization, the training loss can be written as:
\begin{equation*}
    \mathcal{L}(\theta) = - \sum_{t=1}^T \mathbb{E}_q \!\left[ \frac{1}{T-t+1} \sum_{\ell \in \mathcal{M}(x_t)} \log p_\theta\!\left(x_T[\ell] \mid {x}_t \right) \right],
    \label{eq:diffusion-elbo}
\end{equation*}
where $\mathcal{M}(x_t)$ is the set of masked positions in $x_t$. This has a convenient interpretation as the average of masked language modeling (MLM) losses~\cite{devlin-etal-2019-bert} on different masking rates determined by the time step $t$.

\section{Model Training Objectives}

\subsection{Masked Diffusion Models (MDM)}
MDMs are trained by estimating the objective \eqref{eq:diffusion-elbo} with Monte Carlo sampling. 
For $K$ training examples 
$\{{x}_{T,k}\}_{k=1}^K \sim q({x}_T)$, we sample diffusion steps $t_k \sim \operatorname{Unif}(\{1,\dots,T\})$ and then draw a corresponding noisy sequence
${x}_{t,k} \sim q({x}_{t} \mid {x}_{T,k})$ from the forward process. The objective is then estimated by
\begin{equation*}
\widehat{\mathcal{L}}(\theta)
=
- \frac{1}{K}
\sum_{k=1}^K
\frac{1}{T-t_k+1}
\sum_{\ell=1}^L
\log p_\theta\!\left(x_{T,k}^{(\ell)} \mid x_{t_k,k} \right),
\label{eq:diffusion-estimator}
\end{equation*}
where the sum over positions is computed with a single forward pass of the model. We assume the logprob is set to zero for already unmasked tokens, so that only masked positions contribute to the sum. We then take gradient steps against this objective.

\subsection{Semi-Independent and Semi-Causal Diffusion Models (SIDM/SCDM)}
\label{app:sidm-scdm-training}
SIDMs and SCDMs can be trained in essentially the same way as standard MDMs. The only change to training logic is that inputs to the model ($x_{t_k,k}$) must be rearranged as described in \Cref{sec:attn-variants}. We note again that tokens keep their original positional embeddings, so this rearrangement only serves to make the independent/causal attention structures easier to implement.
The quantities
\[p_\theta\!\left(x_{T,k}^{(\ell)} \mid x_{t_k,k} \right)\]
for all masked positions $\ell$ can then be computed in a single forward pass, and training proceeds as usual.

\subsection{AR with Multi-Token Prediction (AR-MTP)}
\label{app:ar-mtp-training}
AR-MTP is trained using an analogous objective to MDMs, with two key differences: (1) we always mask tokens from the right, so each forward state consists of a clean prefix followed by a masked suffix, and (2) we weigh all time steps equally, removing the $\frac{1}{T-t+1}$ normalization. Concretely, at step $t$, the masked set is $\mathcal{M}(x_t) = \{t, t+1, \ldots, L\}$, and the objective is
\begin{equation*}
    \mathcal{L}(\theta) = - \sum_{t=1}^{L} \mathbb{E}_q \!\left[ \sum_{\ell \in \mathcal{M}(x_t)} \log p_\theta\!\left(x_T[\ell] \mid {x}_t \right) \right].
    \label{eq:ar-mtp-objective}
\end{equation*}
Note that if we restrict the inner sum to only the first masked position $\ell = t$, we recover the standard next-token prediction objective of autoregressive language models.
Thus, the additional terms for $\ell > t$ serve as auxiliary prediction losses, encouraging the model to jointly predict multiple future tokens at each position.

\section{Diffusion Inference}
For masked diffusion models, samples are generated by simulating the reverse process for a chosen number of diffusion steps $T$. 

\paragraph{Time conditioning}
We begin by noting an assumption that simplifies inference: the denoising model $p_\theta(x_0 \mid x_t)$ is not explicitly conditioned on the time step $t$. This is a standard choice in works like \citep{sahoo2024simpleeffectivemaskeddiffusion,nie2025largelanguagediffusionmodels,ye2025dream7bdiffusionlarge}. The model simply takes a partially masked state and predicts missing tokens, which requires knowing neither the current step $t$ nor the total steps $T$. This allows us to choose any $T$ during inference and recover the reverse transition probabilities by applying the correct independent probabilities for unmasking each token.

\paragraph{Planned decoding schedules}
\label{app:decoding-general-case}
An equivalent view of the reverse process is to fix the decoding order in advance. 
Let $\mathcal{S} = \{S_1, \dots, S_T\}$ denote a decoding schedule, where each $S_t \subseteq \{1,\dots,L\}$ specifies the set of positions to be decoded at step $t$.
A valid decoding schedule $\{S_t\}_{t=1}^T$ form a partition of $\{1,\dots,L\}$.  Algorithm~\ref{alg:app-mdlm-schedule} summarizes the inference algorithm given diffusion steps $T$ and schedule $\mathcal{S}$. For each step $t$, the model makes a forward pass on $x_t$, and positions in $S_t$ are decoded by sampling from the corresponding predicted distributions, while all other positions remain unchanged.

\begin{algorithm}[t]
\caption{Masked diffusion sampling}
\label{alg:app-mdlm-schedule}
\begin{algorithmic}[1]
\STATE \textbf{Input:} model $p_\theta(x_T[\ell] \mid x)$
\STATE \textbf{Input:} schedule $\mathcal{S}=\{S_1,\dots,S_T\}$
\STATE Initialize $x_T \leftarrow [\textsc{mask}, \dots, \textsc{mask}]$
\FOR{$t = 1, 2, \dots, T$}
    \STATE Initialize $x_{t} \leftarrow x_{t-1}$
    \STATE Compute $p_\theta(x_T[\ell] \mid x_{t-1})$ for all $\ell \in \{1, \dots, L\}$
    \FOR{each $\ell \in S_t$}
        \STATE Sample $x_{t}[\ell] \sim p_\theta(x_T[\ell] \mid x_{t-1})$
    \ENDFOR
\ENDFOR
\STATE \textbf{Return} $x_0$
\end{algorithmic}
\end{algorithm}

Under the standard reverse process, the decoding schedule $\mathcal{S}$ is induced by the independent unmasking decisions at each step, but this can be simulated in advance, or $\mathcal{S}$ can be deliberately chosen.

\section{Latent Token Modulation}
\label{app:latent-token-modulation}
We provide pseudocode in \Cref{alg:latent-token-modulation} for one step of inference using latent token modulation. This replaces lines 4-5 in \Cref{alg:mdlm-schedule}. This algorithm is explained in \Cref{sec:latent-token-modulation}.

\begin{algorithm}[h]
\caption{Latent Token Modulation for SCDMs}
\label{alg:latent-token-modulation}
\vspace{0.5em}
\begin{algorithmic}[1]
\setcounter{ALC@line}{5}
\STATE \textbf{Input:} number of latent tokens $n$
\STATE Sample $Z_t = \operatorname{Perm}(\pi_{>t})[:\min(n, |\pi_{>t}|)]$
\STATE Construct $P_t = \operatorname{concat}(Z_t, [\pi_t])$
\STATE Compute $p_\theta(x[\ell] \mid x_{t-1})$ jointly for $\ell \in P_t$
    \STATE Sample and set $x_{t}[\pi_t] \sim p_\theta(x[\pi_t] \mid x_{t-1})$
\end{algorithmic}
\end{algorithm}

\section{Adaptive Decoding Orders}
\label{app:adaptive-orders}

In this section, we explain how our framework can be made compatible with adaptive decoding orders such as top-probability decoding \citep{kim2025trainworstplanbest}.

\paragraph{Adaptive Decoding Order.} 
Diffusion models can generate tokens in any order. An \emph{adaptive} decoding order means that, at each inference step, we choose which tokens to decode \textit{based on the token distributions} predicted by the model. These methods typically compute a confidence measure for the distribution at each masked position and then decode from the most confident positions first.

\paragraph{Top-prob Decoding.} 
Let $\mathcal{M}_{t} = \{\ell : x_{t-1}[\ell] = \textsc{mask}\}$ denote the set of currently masked positions at step $t$. For each position $\ell \in \mathcal{M}_{t}$, we define the confidence score as the maximum predicted probability over the vocabulary:
\begin{equation}
    c_\ell = \max_{v \in \mathcal{V}} p_\theta(x[\ell] = v \mid x_{t-1}).
\end{equation}
top-prob decoding selects the position with the highest confidence:
\begin{equation}
    \pi_t = \operatorname{argmax}_{\ell \in \mathcal{M}_{t-1}} c_\ell,
\end{equation}
and decodes from this position. Intuitively, we prioritize decoding from positions where the model is most certain. 

\subsection{Adaptive Token Order with SIDM/SCDM}
SIDMs and SCDMs rely on knowing the decoding schedule $\pi$ to construct their inputs (see \Cref{sec:attn-variants}). This might seem problematic when $\pi_t$ is selected on the fly. In this subsection, we describe how to resolve this issue.

At initialization, we sample a random permutation as the tentative schedule $\pi'$, and construct the model input $\tilde{x}_0$ based on this schedule. At each step $t$, we use $P_t = \pi'_{\geq t}$ (all currently masked positions) so that we predict distributions for all masked tokens. top-prob then selects the position $\ell \in P_t$ with highest confidence. After sampling the new token at position $\ell$, we \textit{update} the tentative schedule by moving $\ell$ to position $t$ and shifting the positions between $t$ and $\ell$'s original location. Concretely, if $\pi' = (\ell_1, \dots, \ell_{t-1}, a, b, \ell, c, \dots)$ and top-prob selects $\ell$, the updated schedule becomes $\pi' = (\ell_1, \dots, \ell_{t-1}, \ell, a, b, c, \dots)$. This update maintains the invariant that the first $t$ elements of the schedule are decoded positions, while the remaining elements are masked positions. In the next step, the model input is constructed according to the updated schedule, and this process repeats. \Cref{alg:sidm-topp} provides pseudocode.

\begin{algorithm}[t]
\caption{Adaptive top-prob decoding for SIDM/SCDM}
\label{alg:sidm-topp}
\begin{algorithmic}[1]
\STATE \textbf{Input:} model $p_\theta$, sequence length $L$
\STATE Initialize $x_0 \leftarrow [\textsc{mask}, \dots, \textsc{mask}]$
\STATE Initialize $\pi' \leftarrow \text{random permutation of } (1, \dots, L)$
\FOR{$t = 1, 2, \dots, L$}
    \STATE Construct model input $\tilde{x}_{t-1}$ according to $\pi'$
    \STATE Compute $p_\theta(x[\ell] \mid \tilde{x}_{t-1})$ for all $\ell \in \pi'_{\geq t}$
    \STATE $c_\ell \leftarrow \max_{v \in \mathcal{V}} p_\theta(x[\ell] = v \mid \tilde{x}_{t-1})$ for all $\ell \in \pi'_{\geq t}$
    \STATE $\ell^* \leftarrow \operatorname{argmax}_{\ell \in \pi'_{\geq t}} c_\ell$
    \STATE Sample and set $x_{t}[\ell^*] \sim p_\theta(x[\ell^*] \mid \tilde{x}_{t-1})$
    \STATE Update $\pi'$ by moving $\ell^*$ to position $t$ \COMMENT{Shift intermediate positions}
\ENDFOR
\STATE \textbf{Return} $x_L$
\end{algorithmic}
\end{algorithm}

\subsection{Adaptive Token Order with Latent Token Modulation}
\label{app:adaptive-with-ltm}
The latent token modulation method (\Cref{sec:latent-token-modulation}) introduces another challenge: under adaptive decoding, we do not know which position we will end up decoding, so we cannot rearrange the masked inputs to provide it with a specific number of latent tokens.

To work around this, we restrict the top-prob candidate set to only the next $k$ positions in the tentative schedule, i.e., $\{\pi'_t, \pi'_{t+1}, \ldots, \pi'_{t+k-1}\}$. We rearrange the masked inputs so that these $k$ candidates are preceded by $n$ latent tokens:
\begin{equation}
    [\underbrace{z_1, \ldots, z_n}_{\text{latent tokens}}, \underbrace{\pi'_t, \pi'_{t+1}, \ldots, \pi'_{t+k-1}}_{\text{candidates}}, \ldots]
\end{equation}
Under this arrangement, the $i$-th candidate has access to at least $n$ latent tokens to its left (if there are enough masked tokens in total). We predict distributions for each of the $k$ candidates and apply top-prob to select the position $\ell^*$ with highest confidence. We then proceed as in \Cref{alg:sidm-topp} by moving $\ell^*$ to position $t$ in the schedule and shifting intermediate positions accordingly.

\begin{algorithm}[t]
\caption{Adaptive top-prob decoding with latent token modulation (modifies \Cref{alg:sidm-topp})}
\label{alg:latent-topp}
\vspace{0.5em}
\begin{algorithmic}[1]
\setcounter{ALC@line}{4}
\STATE \textbf{Additional input:} number of latent tokens $n$, number of candidates $k$
\STATE $m \leftarrow |\pi'_{\geq t}|$ \COMMENT{Number of remaining masked positions}
\STATE $C_t \leftarrow \pi'_{t:t+\min(k, m)-1}$ \COMMENT{Candidate set}
\STATE $Z_t \leftarrow \pi'_{t+k:t+k+\min(n, m-k)-1}$ \COMMENT{Latent tokens}
\STATE Construct model input $\tilde{x}_{t-1}$ with masked positions ordered as $[Z_t, C_t, \ldots]$
\STATE Compute $p_\theta(x[\ell] \mid \tilde{x}_{t-1})$ for all $\ell \in C_t$
\STATE $c_\ell \leftarrow \max_{v \in \mathcal{V}} p_\theta(x[\ell] = v \mid \tilde{x}_{t-1})$ for all $\ell \in C_t$
\STATE $\ell^* \leftarrow \operatorname{argmax}_{\ell \in C_t} c_\ell$
\STATE Sample and set $x_{t}[\ell^*] \sim p_\theta(x[\ell^*] \mid \tilde{x}_{t-1})$
\STATE Update $\pi'$ by moving $\ell^*$ to position $t$
\end{algorithmic}
\end{algorithm}

\textbf{For all experiments using top-prob decoding in this paper, we use a candidate set size $k=8$}. For fairness, this applies to models not doing latent token modulation as well (MDM and SIDM). Surprisingly, we find that using such a small candidate set size is already enough to achieve large gains over uniform decoding. For example, \Cref{tab:ar-mtp-reasoning-tasks} includes comparisons of MDM using uniform and top-prob. We also include a version of \Cref{tab:joint-token-ablation} with the added comparison of uniform vs. top-prob below. In both cases, $k=8$ is already able to realize large gains from adaptive decoding. 

\begin{table}
\begin{tabular}{lccc}
\toprule
& Sudoku & Zebra & Countdown \\
\midrule
\multicolumn{4}{l}{Uniform} \\
\midrule
MDM  & 12.1 / 1.00$\times$ & 26.6 / 1.00$\times$ & 17.9 / 1.00$\times$ \\
SIDM & 4.1 / 0.50$\times$ & 14.3 / 0.50$\times$ & 9.4 / 0.50$\times$ \\
\midrule
\multicolumn{4}{l}{top-prob} \\
\midrule
MDM  & 80.2 / 1.00$\times$ & 96.9 / 1.00$\times$ & 27.3 / 1.00$\times$ \\
SIDM & 68.1 / 0.54$\times$ & 71.5 / 0.52$\times$ & 18.4 / 0.60$\times$ \\
\bottomrule
\end{tabular}
\caption{Accuracy (\%) and relative inference cost on reasoning tasks, including both uniform and top-prob decoding. Cost is measured as total tokens processed, reported relative to MDM (1$\times$).}
\label{table1-with-uniform}
\end{table}

\section{Additional Figures}

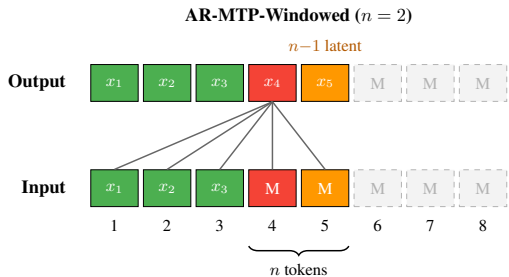
\begin{figure}[h]
    \centering
    \scalebox{0.7}{\begin{tikzpicture}[
    box/.style={draw, minimum width=0.9cm, minimum height=0.7cm, font=\small},
    history/.style={box, fill=historygreen, text=white},
    decode/.style={box, fill=decodered, text=white},
    latent/.style={box, fill=latentorange, text=white},
    ignored/.style={box, fill=ignoredgray, text=black!60},
    omitted/.style={box, draw=black!30, dashed, fill=ignoredgray!40, text=black!30},
    poslabel/.style={font=\footnotesize},
    attention/.style={thick, black!60},
    rowlabel/.style={font=\bfseries, anchor=east},
    titlelabel/.style={font=\bfseries, anchor=south},
]

\def\boxsep{1.0}
\def\rowsep{2.0}

\node[titlelabel] at ({4.5*\boxsep}, 1.0) {AR-MTP-Windowed ($n=2$)};

\def\yout{0}

\node[history] (out1) at ({1*\boxsep}, \yout) {$x_1$};
\node[history] (out2) at ({2*\boxsep}, \yout) {$x_2$};
\node[history] (out3) at ({3*\boxsep}, \yout) {$x_3$};
\node[decode] (out4) at ({4*\boxsep}, \yout) {$x_4$};
\node[latent] (out5) at ({5*\boxsep}, \yout) {$x_5$};
\node[omitted] (out6) at ({6*\boxsep}, \yout) {M};
\node[omitted] (out7) at ({7*\boxsep}, \yout) {M};
\node[omitted] (out8) at ({8*\boxsep}, \yout) {M};

\node[rowlabel] at (0.2, \yout) {Output};

\def\yin{-\rowsep}

\node[history] (in1) at ({1*\boxsep}, \yin) {$x_1$};
\node[history] (in2) at ({2*\boxsep}, \yin) {$x_2$};
\node[history] (in3) at ({3*\boxsep}, \yin) {$x_3$};
\node[decode] (in4) at ({4*\boxsep}, \yin) {M};
\node[latent] (in5) at ({5*\boxsep}, \yin) {M};
\node[omitted] (in6) at ({6*\boxsep}, \yin) {M};
\node[omitted] (in7) at ({7*\boxsep}, \yin) {M};
\node[omitted] (in8) at ({8*\boxsep}, \yin) {M};

\node[rowlabel] at (0.2, \yin) {Input};

\foreach \i in {1,...,8} {
    \node[poslabel] at ({\i*\boxsep}, \yin-0.7) {\i};
}

\draw[decorate, decoration={brace, amplitude=5pt, mirror}, thick] 
    ({4*\boxsep-0.45}, \yin-1.1) -- ({5*\boxsep+0.45}, \yin-1.1)
    node[midway, below=6pt, font=\footnotesize] {$n$ tokens};

\node[font=\footnotesize, text=orange!70!black, above=3pt] at (out5.north) {$n{-}1$ latent};

\draw[attention] (in1.north) -- (out4.south);
\draw[attention] (in2.north) -- (out4.south);
\draw[attention] (in3.north) -- (out4.south);
\draw[attention] (in4.north) -- (out4.south);
\draw[attention] (in5.north) -- (out4.south);

\end{tikzpicture}}
    \caption{Windowed AR-MTP architecture. The model jointly predicts a window of $n$ masked tokens; all subsequent tokens are ignored.}
    \label{fig:ar-mtp-win}
\end{figure}

\section{Experimental Details}
\subsection{Models}
\label{app:model-sizes}
All models considered are based on the transformer architecture. For each task, we use the same model size across methods. We consider these model configurations:

\begin{table}[h]
\centering
\caption{Model configurations and parameter counts. Parameter counts are computed for a vocabulary size of 32, which is representative for synthetic tasks.}
\label{tab:model-configs}
\begin{tabular}{lcccccc}
\toprule
\textbf{Model} & \textbf{Hidden} & \textbf{Heads} & \textbf{Layers} & \textbf{Params} \\
\midrule
tiny  & 384  & 12 & 3  & 6.4M \\
mini       & 512  & 8  & 6  & 21.5M \\
sminy      & 768  & 12 & 6  & 46.4M \\
small      & 768  & 12 & 12 & 92.4M \\
\bottomrule
\end{tabular}
\end{table}

\subsection{Datasets}
\label{app:datasets}
Table~\ref{tab:datasets} summarizes the key statistics of each dataset.

\begin{table}[h]
\centering
\caption{Dataset statistics and training configurations.}
\label{tab:datasets}
\begin{tabular}{lrrr}
\toprule
\textbf{Dataset} & \textbf{Train Size} & \textbf{Seq. Length} & \textbf{Vocab} \\
\midrule
Sudoku-Gen & 100K & 128 & 11 \\
Sudoku-Puzzle & 1.8M & 192 & 14 \\
Zebra & 1.5M & 384 & 23 \\
Countdown & 500K & 64 & 20 \\
Sudoku-Large & 100K & 1536 & 38 \\
OpenWebText & - & 1024 & 50K \\
\bottomrule
\end{tabular}
\end{table}

\paragraph{Sudoku-Generative.} We generate valid 9$\times$9 Sudoku solutions using a constraint-based solver. Each example consists of 81 digit tokens (1--9) representing the complete board in row-major order. The vocabulary includes digits 1--9, mask, and special tokens. The model is trained to generate valid boards unconditionally.

\paragraph{Sudoku-Puzzle.} This dataset pairs Sudoku puzzles with their solutions. Each sequence contains the puzzle (with empty cells marked), a separator, and the solution: \texttt{[BOS] puzzle(81) [SEP] solution(81) [EOS]}. The vocabulary adds a separator token to distinguish puzzle from solution.

\paragraph{Zebra.} Einstein's riddle-style logic puzzles where the model must deduce attribute assignments across multiple houses from a set of symbolic clues. Each puzzle consists of clues encoded as tokens followed by the solution grid. The vocabulary includes digits, logical operators (=, !=), positional relations (left-of, immediate-left, etc.), and clue markers.

\paragraph{Countdown.} Arithmetic chain puzzles. Given 5 input numbers and a target, the model must produce a sequence of arithmetic operations reaching the target. Each step has the form \texttt{a op b = c}, with operations connected by commas. The vocabulary contains digits (0--9), operators (+, -, *, /), and delimiters.

\paragraph{Sudoku-Large.} A scaled-up variant of Sudoku using base-6, resulting in a 36$\times$36 grid with 6$\times$6 blocks. Each cell contains a digit from 1--36, and the standard Sudoku constraints apply: each row, column, and block must contain all 36 digits exactly once. The vocabulary includes 38 tokens covering the 36 digit values plus special tokens. With 1,296 cells, this task tests the model's ability to maintain long-range dependencies and satisfy numerous interacting constraints simultaneously.

\textbf{OpenWebText.} A natural language dataset \citep{Gokaslan2019OpenWeb} replicating the training data used for GPT-2. We do not train on this dataset ourselves, instead using the model checkpoint released by \citet{sahoo2025esotericlanguagemodels}.

\subsection{Training}
Table~\ref{tab:data-model-size-epochs} shows the model size and number of epochs trained on each dataset. 

\begin{table}[h]
\centering
\caption{}
\label{tab:data-model-size-epochs}
\begin{tabular}{lcc}
\toprule
\textbf{Task} & \textbf{Model} & \textbf{Epochs} \\
\midrule
Sudoku-Gen & sminy & 100 \\
Sudoku-Puzzle & mini & 20 \\
Zebra & mini & 50 \\
Countdown & tiny & 300 \\
Sudoku-Large & sminy & 50 \\
OpenWebText & small & - \\
\bottomrule
\end{tabular}
\end{table}

For OpenWebText, we use the checkpoint released by \citet{sahoo2025esotericlanguagemodels}. We train all models for the indicated number of epochs and choose the best checkpoint by validation loss. For countdown, we only train the standard ARM for 40 epochs as we found it already converges at this point and begins to overfit. This is consistent with the training configs in \citep{ye2025autoregressiondiscretediffusioncomplex}.

\subsection{Inference}
When evaluating diffusion models, we always use number of diffusion steps equal to sequence length ($T=L$), and fix the algorithm to generate one token per step (\Cref{alg:mdlm-schedule}) instead of stochastically unmasking.

\paragraph{Decoding Method} When not specifically stated, we always use top-prob decoding for reasoning tasks (\Cref{fig:latent-tokens-four-datasets}, \Cref{tab:joint-token-ablation}) and use uniform decoding for sequence modeling tasks (\Cref{fig:latent-tokens-owt}, \Cref{fig:sudoku-large-plot}). For top-prob decoding, we always use candidate set size $k=8$ -- this is explained in App.~\ref{app:adaptive-with-ltm}.

Specifically for the Countdown task, we follow \citet{ye2025autoregressiondiscretediffusioncomplex} and decode tokens greedily (i.e., we do not sample from the predicted token distribution but take the argmax). We find that performance is very poor on this task without this configuration. 

\subsection{Sudoku-Large Task Metric}
\label{app:sdk-large-task-loss}
For the Sudoku-Large task, we compute a soft constraint loss that measures how close a generated board is to satisfying Sudoku rules, providing a more granular metric than binary validity. For each constraint unit $u$ (row, column, or block), we count the number of unique valid digits present and compute the unit loss as $\ell_u = 1 - |\text{unique}(u) \cap \mathcal{V}| / n$, where $\mathcal{V} = {0, \ldots, n-1}$ is the set of valid digits and $n = \text{base}^2$ is the grid side length. The total constraint loss is the sum over all $3n$ units: $\mathcal{L} = \sum_{u \in \text{rows} \cup \text{cols} \cup \text{blocks}} \ell_u$. For a base-6 Sudoku ($n=36$), this yields 108 constraint units. A perfectly valid board achieves $\mathcal{L} = 0$, while the worst case (e.g., all identical digits) yields $\mathcal{L} = 3(n-1) = 105$. Invalid tokens outside $\mathcal{V}$ do not reduce the loss, effectively penalizing them at least as much as duplicates.

\section{Discussion}
\subsection{Analogy to Latent Reasoning}
\label{app:analogy-latent-reasoning}
Consider the task of generating an answer $A$ given a question $Q$.
It is well known that directly modeling $P(A \mid Q)$ performs poorly on complex reasoning tasks.
Instead, substantially stronger performance can be obtained by modeling
\[
P(A, R \mid Q) = P(R \mid Q) P(A \mid R, Q),
\]
where $R$ represents an explicit reasoning process, such as a chain of thought \citep{wei2023chainofthoughtpromptingelicitsreasoning}.
This observation suggests that while $P(A \mid Q)$ may be difficult to learn directly, there exists a latent variable $R$
for which both $P(R \mid Q)$ and $P(A \mid R, Q)$ are more tractable, enabling recovery of $P(A \mid Q)$ via marginalization.

Recent work further suggests that such reasoning processes need not to be discrete tokens.
Instead, models may learn latent, continuous thought representations that implicitly encode
reasoning trajectories \citep{hao2025traininglargelanguagemodels}.
Probing analyses indicate that these latent thoughts simultaneously encode multiple plausible reasoning paths, i.e., many values of $R$.
The model can then implicitly marginalize over reasoning paths to produce predictions for $P(A \mid Q)$.

In masked diffusion models, latent tokens can be viewed as playing an analogous role to the reasoning variable~$R$. In a denoising step $t$, the model conditions on $x[\pi_{<t}]$ and predicts target token $x[\pi_t]$, while jointly computing but not decoding latent tokens $x[\pi_{>t}]$.  
Predicting $x[\pi_t]$ in isolation may be difficult when these tokens depend on constraints involving many other unknown tokens (e.g., filling a cell in a Sudoku grid). However, when the model jointly computes distributions $p_\theta(x_0[\pi_{>t}] \mid x_t)$, the hidden representations associated with these positions can participate in the prediction. This allows the model to exploit information about plausible values of other unknown tokens when generating the next token, in a manner analogous to the reasoning strategies discussed above.

\section{Additional Related Works}
\textbf{Diffusion Inference Optimization.}
Unlike ARMs, MDMs do not naturally support KV caching. Several works address this limitation by introducing approximate caching schemes \citep{wu2025fastdllmtrainingfreeaccelerationdiffusion,ma2025dkvcachecachediffusionlanguage} or modifying the architecture to enable exact caching \citep{sahoo2025esotericlanguagemodels}. In this work, we take a complementary approach: controlling the amount of joint computation as a mechanism to trade off inference cost and sample quality. Our semi-causal diffusion model builds on \citet{sahoo2025esotericlanguagemodels}, who introduced this attention pattern for efficient inference.

\paragraph{Multi-token Prediction.}
\citet{gloeckle2024betterfasterlarge} and \citet{deepseekai2025deepseekv3technicalreport} train ARMs with multi-token prediction objectives to provide additional supervision and enable speculative decoding. Our AR-MTP uses a similar architecture, but due to our bidirectional design, the extra tokens serve not only as additional supervision but also as additional inference compute, since they always participate in predicting the next token.


\end{document}